\documentclass[lettersize,journal]{IEEEtran}
\usepackage{amsmath,amsfonts,amssymb}
\usepackage{algorithmic}
\usepackage{algorithm}
\usepackage{array}
\usepackage[caption=false,font=normalsize,labelfont=sf,textfont=sf]{subfig}
\usepackage{textcomp}
\usepackage{stfloats}
\usepackage{url}
\usepackage{verbatim}
\usepackage{graphicx}
\usepackage{cite}
\usepackage[misc]{ifsym}

\usepackage{color}
\usepackage{xcolor}
\usepackage{import}
\graphicspath{{images/}}
\usepackage{bm}
\usepackage{svg}
\usepackage[flushleft]{threeparttable}
\usepackage{siunitx} 
\usepackage{booktabs}
\usepackage{amsmath}
\usepackage[hidelinks]{hyperref}
\usepackage{cleveref}
\usepackage{adjustbox} 
\Crefname{figure}{Fig.}{Fig.}
\Crefname{section}{Sec.}{Sec.}

\DeclareMathOperator*{\argmin}{arg\,min}

\definecolor{Reviewer1}{rgb}{0,0,0.6}
\definecolor{Reviewer2}{rgb}{0.8,0.6,0}
\definecolor{Reviewer3}{rgb}{0,0.6,0}
\definecolor{Reviewer4}{rgb}{0.8,0,0.6}
\definecolor{Reviewer5}{rgb}{0,0.6,0.8}
\definecolor{Reviewer6}{rgb}{0.6,0,0}
\definecolor{AE}{rgb}{0.6,0,0}

\begin{document}


\title{Multi-tap Resistive Sensing and FEM Modeling enables Shape and Force Estimation in Soft Robots}


\author{Sizhe Tian$^{1,\dagger}$, Barnabas Gavin Cangan$^{1,\dagger}$, Stefan Escaida Navarro$^{2,3}$,\\ Artem Beger$^{4}$,
Christian Duriez$^{3}$, and Robert K. Katzschmann$^{1,*}$

\thanks{$\dagger$ denotes equal contribution}
\thanks{$^{1}$Soft Robotics Lab, IRIS, D-MAVT, ETH Zurich, Switzerland
        {\texttt{sizhe.tian@alumni.ethz.ch,bcangan@ethz.ch,rkk@ethz.ch}}}%
\thanks{$^{2}$Instituto de Ciencias de la Ingeniería, Universidad de O'Higgins, Chile
        {\texttt{ stefan.escaida@uoh.cl}}}%
\thanks{$^{3}$INRIA – Lille – Nord-Europe, Lille University, France
        {\texttt{ christian.duriez@inria.fr }}}%

\thanks{$^{4}$Festo SE \& Co. KG, Esslingen am Neckar, Germany
        {\texttt{ artem.beger@festo.com}}}%
\thanks{$*$ Corresponding author: \href{mailto:rkk@ethz.ch}{\tt rkk@ethz.ch}}

}




\maketitle
\thispagestyle{empty}

\begin{abstract}
We address the challenge of reliable and accurate proprioception in soft robots, specifically soft robots with tight packaging constraints and relying only on internally embedded sensors. While various sensing approaches with single sensors have commonly been tried using a constant curvature assumption, we look into sensing local deformations at multiple sensor locations.
%
In our approach, we multi-tap an off-the-shelf resistive sensor by creating multiple electrical connections onto the resistive layer of the sensor, and we insert the sensor into a soft body. This modification allows us to measure changes in resistance at multiple segments throughout the length of the sensor, providing improved resolution of local deformations in the soft body. These measurements inform a model based on a finite element method (FEM) that estimates the shape of the soft body and the magnitude of an external force acting at a known arbitrary location.
%
Our model-based approach estimates soft body deformation with approximately 3\% average relative error while taking into account internal fluidic actuation. Our estimate of external force disturbance has an 11\% relative error within a range of 0 to 5 N.
%
For instance, the combined sensing and modeling approach can be integrated into soft manipulation platforms to enable features such as identifying the shape and material properties of an object being grasped. Such manipulators can benefit from the inherent softness and compliance while being fully proprioceptive, relying only on embedded sensing and not on external systems such as motion capture. Such proprioception is essential for the deployment of soft robots in real-world scenarios.

\end{abstract}


\section{Introduction}
\label{sec:Intro}
\label{subsec:Proprioception}

\begin{figure}[ht]
    \centering

    \fontsize{7}{9}\selectfont
    \def\svgwidth{0.98\linewidth}
    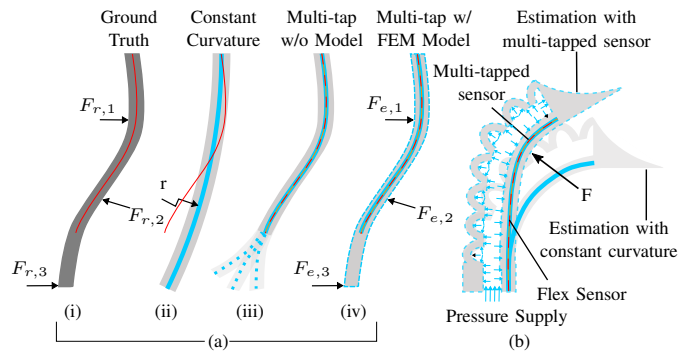
    
    \setlength{\belowcaptionskip}{-4pt}
    \caption{\textbf{Advantages of a multi-tapped flex sensor with continuum model.}
    \textbf{(a)} Reconstruction results of the shape of a flexible strip using different sensing and modeling strategies. The \textbf{\textcolor{darkgray}{dark gray}} stripes represent true shape of the flexible strips and \textbf{\textcolor{lightgray}{light gray}} stripes their estimated shape. The \textbf{\textcolor{red}{red}} lines are the flex sensors, the overlaid \textbf{\textcolor{orange}{orange}} dashed lines indicate multi-tapping, and the \textbf{\textcolor{cyan}{cyan}} lines represent shape predictions. Dashed \textbf{\textcolor{cyan}{cyan}} outline indicates shape estimated using an FEM-based model.
    (i) Actual shape of a flexible strip with an embedded flex sensor bent under known forces $\text{F}_{r, i}$.
    (ii) Given a single flex sensor reading, model with constant-curvature assumption estimates radius of curvature $r$.
    (iii) Multi-tapped flex sensor without our modeling approach cannot estimate the shape of the flexible strip beyond the active length of the flex sensor (\textbf{\textcolor{cyan}{cyan}} dotted lines).
    (iv) Multi-tapped sensing with a model allows the full shape and forces at known points $\text{F}_{e, i}$ to be estimated. 
    \textbf{(b)} For a soft pressurized finger, shape estimation results using a single flex sensor with constant-curvature assumption compared to our multi-tapped sensing strategy with continuum model.
    }
    \label{fig:sensor_pro}
\end{figure}
Soft robots are robotic systems with purposefully designed compliant elements embedded into their mechanical structure.
Thereby, soft robots can passively adapt within unstructured environments and are safer when interacting with humans~\cite{della_santina_soft_2020}.
Despite these benefits, instrumenting soft robots to be proprioceptive, \textit{i.e.}, correctly perceiving their own shape \cite{balasubramanian2014human} in contexts without and with external interactions, remains an open challenge.


The shape of a rigid robot is fully defined by the configuration of its fixed number of joints independent of interactions with one or more loads acting on the system~\cite{martin_proprioceptive_2022}. In a soft robot, however, each such interaction creates deformations associated with it. Proprioception in soft robotics, therefore, requires being able to sense these deformations to reason about the state of the system and the load\cite{balasubramanian2014human}   . 
We consider a suitable performance in proprioception when a soft robot can sense its own configuration within an acceptable error margin, which in this work, we define to be $<$5$\%$ of the characteristic length~\cite{martin_proprioceptive_2022,baaij_learning_2023}.
Furthermore, to be able to estimate load, the approach needs to generalize to the full range of control inputs and states reachable by the soft robot's internal actuation. Thereby, we postulate that high-performance proprioception in soft robotics has to fulfill the following three requirements:
(1) high-resolution sensing of local deformations,
(2) a model of the soft device, including its inlaid sensor, and
(3) a model for the device actuation.

In \Cref{sec:Related Work}, we shall further discuss these requirements in the context of placing current approaches in soft robot proprioception within this framework and relating them to each other using \Cref{fig:sensor_pro}. The figure highlights the key aspects of our approach and how it relates to other state-of-the-art methods.
Following that, \Cref{sec:FlexSensor} introduces the multi-tapped flex sensor technology and how the shape descriptors (position and angular displacements) are calculated from resistance values.
\Cref{sec:devices} discusses the flexible strip and soft finger used to validate the proposed proprioceptive approach.
\Cref{sec:ModelingTechniques} describes the finite element model defined in Simulation Open Frameware Architecture (SOFA)\footnote{\url{https://www.sofa-framework.org}}. The experimental validation is presented in \Cref{sec:Experiments}.
Finally, \Cref{sec:Conclusions} summarizes our contributions and provides an outlook to future developments.

\section{State-of-the-Art and Our Contribution}
\label{sec:Related Work}

\subsection{Shape Proprioception with Single Curvature Measurement}
Several works have already proposed an integrated design and fabrication approach to bring proprioceptive sensors into soft robots~\cite{gu20233d}.
Embedded sensors with various sensing modalities can achieve proprioceptive sensing in a soft gripper towards closed-loop manipulation of the object ~\cite{truby_soft_2018,truby2019soft}.
Also, commercially available resistive flex sensors in a soft gripper can detect the object being grasped~\cite{homberg_robust_2019}.
These approaches, however, fall short in meeting the high-resolution sensing requirement we laid out in \Cref{subsec:Proprioception}; they use a single measurement with a constant-curvature assumption as shown in \Cref{fig:sensor_pro}~(ii), \textit{i.e.}, the sensing resolution is low and local deformations are not captured. Also capacitive flex sensors\footnote{BendLabs, Inc.} can output coarse shape estimations when local curvatures along the sensor have different signs and therefore cancel each other out~\cite{toshimitsu_sopra_2021}.

\subsection{Data-driven Force Estimation with Proprioception}
Force estimation using proprioceptive sensing can be achieved by either adding force sensors at the tip of the robot~\cite{park_design_2012, morrow_improving_2016} or building a robot model based on the Piecewise Constant Curvature (PCC)~\cite{ c_della_santina_dynamic_2018, wang_contact_2021} or Cosserat-rod theory~\cite{f_renda_dynamic_2014, c_armanini_discrete_2021}.
Data-driven approaches have also been proposed to address the challenge of force proprioception~\cite{elgeneidy_bending_2018,thuruthel_soft_2019}.
Recurrent neural networks (RNN) can be used to learn the sensor characteristics of three resistive strain sensors in a pneumatically actuated soft robotic finger and to thereby estimate the external force at the tip of a soft finger~\cite{thuruthel_soft_2019}. 
This learning approach~\cite{thuruthel_soft_2019} allows for proprioception with relatively high spatial resolution and helps with issues like hysteresis;  
however, such approaches fall short of meeting requirements (2) and (3), with an outcome similar to what is illustrated in \Cref{fig:sensor_pro}~(iii). Since it is purely data-driven rather than based on the known mechanical properties of the device, the lack of a mechanical model causes the proprioception not to extend to areas not directly observable by the sensor.

\subsection{Model-based Proprioception}
Modes et al.~\cite{modes2020shape} present a model based on the Cosserat rod theory to determine the shape of a sensor array based on accurate longitudinal strain measurements and incorporating bending, twisting, and elongation. 
Tapia et al.~\cite{tapia2020makesense} present an approach for optimal sensorization of soft robots using stretch sensors to reconstruct their shape using an FEM-based mechanical model and a model for resistive type sensors. In their designs, the sensors cover the whole length of the devices, which allows for finding the deformations due to actuation efforts and external forces. 
However, the focus is on shape reconstruction, and they stop short of explicitly quantifying force estimation performance or extrapolating shape beyond sensorized regions using their approach.

In our previous work, we introduced a multi-modal method to estimate shape and forces based using FEM~\cite{escaida_navarro_model-based_2020}. Capacitive touch sensing is used to determine the force location on the devices, and air-flow/pressure sensors are used to solve an optimization problem that yields the force magnitudes that best explain the observed changes in sensor values (volume/pressure). This work was followed up by an approach featuring orientation-sensing (one reading per segment) for a fluidically actuated soft arm~\cite{cangan2022model}. Here, attention is paid to modeling the effect of actuation and the fiber-reinforcement constraining the deformation of the pressurized chambers. Koehler et al.\ proposed a model-based sensing method with model-calibrated embedded sensors by solving the inverse problem to find the optimal shape that will lead to the sensor readings~\cite{koehler2020model}. However, using high-resolution resistive shape sensing for proprioception (shape and force) has not yet been explored in this framework.

\subsection{Our Contribution}
In this work, we use an inexpensive, commercially available 95 mm Spectra Symbol Resistive Flex Sensor.
This sensor allows multi-tapping, i.e to make measurements at multiple exposed electrical connections, along the resistive strip.
With the additional measurements, the sensor is able to capture local curvature changes for high-resolution shape measurements. 
The multi-tapped sensor, along with the solid mechanics of the device and the pressure-based actuation are modeled in an FEM-based framework. The modeling allows us to generalize the shape and force estimation to parts of the soft body not directly observable by the sensor.

We address the requirements from \Cref{subsec:Proprioception} by contributing (1) the creation of a multi-tapped resistive shape sensor using an off-the-shelf sensor and embedding it in two types of soft devices, (2) the modeling of soft devices with embedded sensors within a finite element modeling framework (SOFA), and (3) modeling within this framework the actuation by pressurization of internal chambers. 
We embed the shape sensors for validation in two devices, a flexible strip without actuation (\Cref{fig:sensor_pro}~(iv)) and a soft finger with actuation (\Cref{fig:sensor_pro}~(b)).
We experimentally validate the proposed approach and show that accurate shape $(\text{avg. error }<3\%)$ and force estimation $(\text{avg. error }<12\%)$ is possible with the devices (shape for the flex strip and shape and force for the soft finger). 
However, certain uncaptured nonlinearities still occur, which introduced errors in our system. The model also needs further optimization to run fast enough for real-time estimation.

\section{System Description}
\subsection{Multi-tap Flex Sensor}
\label{sec:FlexSensor}
\subsubsection{Design and Fabrication}
\begin{figure}[ht]
    \centering

    \fontsize{7}{9}\selectfont
    \def\svgwidth{0.95\linewidth}
    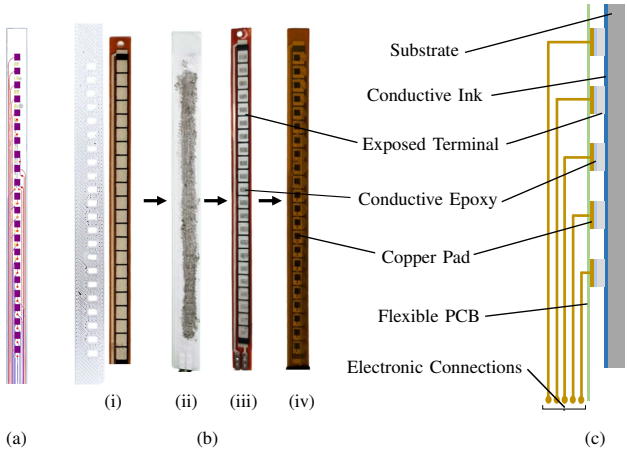

    \caption{\textbf{Fabrication and design of the multi-tap flex sensor.} (a) Diagram of the flexible interface PCB (b) Fabrication process of the multi-tap flex sensor. (i) The flex sensor and a 3D-printed mask. (ii) Apply conductive epoxy to the flex sensor using the mask. (iii) Conductive epoxy applied on the exposed contact points is visible after the removal of the mask. (iv) Flexible interfacing PCB is attached to the sensor and held together by the conductive epoxy. (c) Structure of the multi-tap flex sensor.
    }
    \label{fig:multi_tap_sensor_fab}
\end{figure}

The \emph{Spectra Symbol} flex-sensor consists of a thin strip of flexible conductive material coated onto a flexible but non-stretchable substrate. When the sensor is bent, the conductive ink is stretched, and its electrical resistivity changes. The resulting change in resistance can be measured to determine the radius of curvature. We multi-tap this sensor to extend its capability to measure local deformations along its length using a simple flexible printed circuit board (PCB) designed to route individual tapping points to the measurement circuit. A diagram of the PCB layout is shown in Fig. \ref{fig:multi_tap_sensor_fab}. The flexible interfacing PCB is glued to the flex sensor on the side with the exposed pads to make electrical contact using an epoxy glue\footnote{CircuitWorks 2-part Conductive Epoxy CW2460}.
There are four simple steps to fabricating the multi-tap flex sensor as shown in \Cref{fig:multi_tap_sensor_fab}. 
Firstly, prepare the necessary materials: the flex sensor, flexible PCB, epoxy glue, and a 3-D printed mask (0.2 mm thick) with windows that match the exposed pads on the sensor.
Secondly, place the mask on the flex sensor, align the windows to the tapping points, and apply the epoxy glue on the mask.
Thirdly, remove the mask and check that the epoxy glue is distributed evenly on all tapping points without any bridging between adjacent contact pads.
Lastly, place the flexible PCB on the sensor and wait for the epoxy glue to cure.
The fabrication process of the multi-tapped flex-sensor is shown in \Cref{fig:multi_tap_sensor_fab}. This design satisfies requirement (1) on high-resolution shape sensing, established in \Cref{sec:Intro}.

\subsubsection{Mapping Resistance to Shape}
\label{subsec:MappingRtoShape}

While multi-tapping allows us to measure changes in resistance across the sensor with better spatial resolution, the mapping from these resistance measurements to the shape of the sensor is not trivial to model. The segments across which we measure resistance are part of the same substrate, and therefore their resistance values are correlated. Local deformation at one end of the sensor still affects the resistance across a segment on the other end, albeit to a lesser extent. Hence, we used a neural network to learn this mapping from resistance measurements to shape vectors.

\begin{figure}[ht]
    \centering

    \fontsize{7}{9}\selectfont
    \def\svgwidth{0.65\linewidth}
    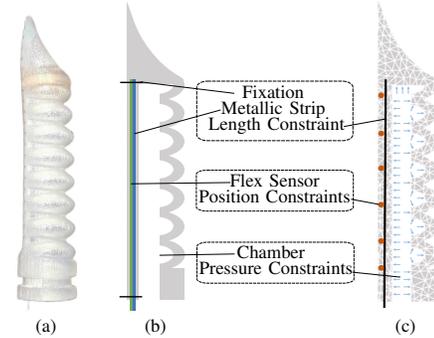

    \caption{\textbf{Structure and computational model of the soft finger.} (a) Structure of soft finger. (b) Design of the soft robot finger. (c) Diagram of the FEM model for the soft finger. Boxes represent correspondence in design and the FEM model.
    }
    \label{fig:sensor_finger_diagram}
\end{figure}

The shape vectors, in our case, are the displacement (translational and rotational) of a set of points matching the placement of the sensor within the soft body. Corresponding points at the same locations are established on the FEM mesh in simulation. A component in SOFA named the \emph{pose effector} (see \Cref{subsec:ForceOptimization}) serves to apply measured displacements as constraints to an optimization problem, allowing to determine the efforts (internal actuation, external forces) that lead to the sensed shape of the flexible strip or the soft finger (see \Cref{fig:sensor_finger_diagram}).

Thus, the shape model of the resistive flex sensor can be expressed as
\begin{equation}
    \bm{S}= \mathcal{F}(\bm{R})
\end{equation}
where $\bm{S}$ is an array of poses that denotes the shape of the soft robotic device where the sensor is embedded at any time. $\bm{R}$ denotes the vector of resistance measurements corresponding to the sensor segments, and $\mathcal{F}$ is the mapping between the two we learn from observations
Note that depending on the scenario, $\bm{S}$ can contain only the translational or rotational portion of the poses. Furthermore, not all degrees of freedom need to be active; for instance, one of the dimensions could be ignored because deformation is only described in a plane. We provide the corresponding details on experimental validation in \Cref{sec:Experiments}. 

\subsubsection{Nominal Shape Vector}
\label{subsec:NominalShapeArray}

For neural network training, ground truth is established from visual markers whose pose is captured with a camera. The visual markers are attached according to features on the devices that are foreseen during their fabrication (see \Cref{fig:demo_sys}). Therefore, there is a \emph{nominal position} for each marker that we can also establish in simulation. 

Nonetheless, discrepancies due to fabrication tolerances are still present, which prevents the direct use of the visual marker to feed the simulation. 
It becomes necessary to establish an additional (barycentric) mapping that converts the measured marker position to its nominal counterpart. 

The nominal shape vector $\bm{S}_{nom}$ is gathered from the FEM model with the following procedure. One point is assumed to be exactly coincident in the camera frame and the simulation, which is a fixed point on the base of the device. This point was chosen as it would be a location least sensitive to fabrication tolerances.
 Marker positions are captured relative to this base point and initialized directly in the geometric model. We denote the initial marker positions as $\bm{p}_0$. Then we select a set of evenly spaced predetermined locations on the geometric model, which we denote as $S_0$. These points correspond to the position of the sensor within the device for the shape vector $\bm{S}_{nom}$. The marker positions and shape vector are connected to the mesh via barycentric mapping. Therefore, as the device deforms, the relative poses between the markers and the nominal poses are kept coherent through the mechanical model. The nominal shape vector $S_{nom}$ is the output of the sensor regression and is fed to the simulation.  


\subsection{Flexible Strip and Soft Finger}
\label{sec:devices}

\begin{figure}[ht]
 
    \fontsize{7}{9}\selectfont
    \def\svgwidth{\linewidth}
    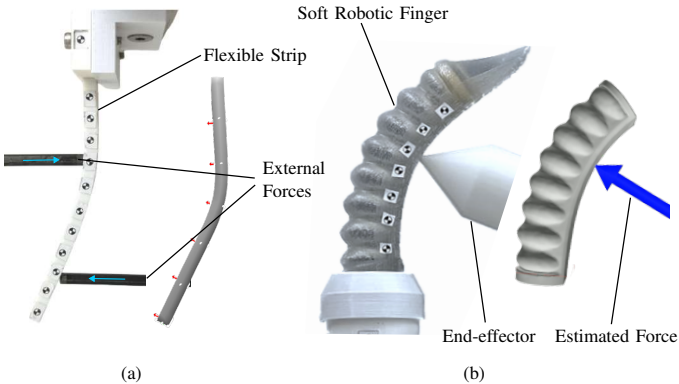

    \caption{\textbf{Multi-tap sensor and soft devices tested with their FEM models.}
    (a) Flexible strip deformed under external force (left) and its simulation in the SOFA (right).
    (b) Pressure-actuated soft finger deformed under external force using a force-controlled robotic arm (left) and the finger's simulation in SOFA (right).
    }
    \label{fig:demo_sys}
\end{figure}

The proposed concept is evaluated using the hardware and systems shown in \Cref{fig:demo_sys}. 
Overall, the system is based on a multi-tapped resistive flex sensor.
The sensor was integrated into different devices to investigate its sensing capability.
Additionally, measurements of visual markers placed on the soft body serve as ground truths in the sensor model training.
With the trained sensor model, the requirement (1) from \Cref{sec:Intro} can be fulfilled.

To test the multi-tapped flex sensor's capabilities, we embedded it into a 3D-printed flexible thermoplastic polyurethane strip (Ultimaker TPU 95A) with a narrow slit in the middle designed to insert the multi-tapped flex sensor. 
Furthermore, the multi-tapped sensor is embedded in a soft finger 3D-printed in an elastomer material which is part of a soft gripper designed by Festo SE \&
Co. KG. The finger comprises a chamber for pneumatic actuation and an embedded metallic strip to act as a constraint layer for bending. However, the finger is relatively stiff, and the shape-sensing capabilities of the multi-tapped sensor cannot be fully investigated with this device. Therefore, we introduced the flexible strip to test shape-sensing. 

We used the SOFA framework and its Soft Robots plugin \cite{coevoet_software_2017} to build a model based on the finite element method. The simulated model running in SOFA is updated according to the sensor values, yielding a shape estimate (see \Cref{fig:demo_sys}~(a) and (b)). 
After calibration using a force-torque sensor, SOFA is able to represent both the effect of known pressurization as well as unknown external forces (see \Cref{fig:demo_sys}~(b)) without the need for any additional sensing modalities.
Thereby the simulation model satisfies requirements (2) and (3) from \Cref{sec:Intro}.

\section{Modeling Framework}

\label{sec:ModelingTechniques}

\subsection{Equations of Quasi-Static Motion}
\label{subsec:EquationsOfStaticEquilibrium}

We use an FEM mesh which yields the internal elastic forces $\mathbb{F}(\bm{q})$, given that the nodes are at positions $\bm{q}$. We choose Hooke's law to describe the relation of force and deformation, i.\,e.\ the material properties are characterized by Young's modulus and Poisson's ratio.
At each step $i$ of the simulation, we compute a linearization of the internal forces as follows:
\begin{equation}
\label{eq:linearization}
\mathbb{F}(\bm{q_{i}}) \approx \mathbb{F}(\bm{q_{i-1}}) + K(\bm{q_{i-1}})d\bm{q_i}
\end{equation}
where $d\bm{q_i = q_{i}-q_{i-1}}$ is the node displacement and $K=\frac{\partial \mathbb{F}(\bm{q_{i-1}})}{\partial \bm{q}}$ is the tangential stiffness matrix for the current node positions $\bm{q}$. Now, we add external forces to the equation for a complete picture:
\begin{equation}
\label{eq:equilibrium}
0 = -K(\bm{q_{i-1}})d\bm{q} + \mathbb{P} - \mathbb{F}(\bm{q_{i-1}}) + H^{T}\bm{\lambda}
\end{equation}
$H^{T}\bm{\lambda}$ is a vector that gathers boundary forces, such as contacts or external controlled inputs. The size of $\bm{\lambda}$ is equal to the number of rows in $H$ and to the number of actuators (contact forces, cables, etc.). $\mathbb{P}$ represents known external forces, such as gravity. Then, (\ref{eq:equilibrium}) is solved under the assumption of a static equilibrium at each timestep, i.\,e.\ the left-hand side is 0, as the effects of acceleration can be neglected in our setting. This delivers a motion that is a succession of quasi-static states.

\subsection{Force, Pressure, and Length Constraints}
\label{subsec:Constraints}
%
In our FEM-based simulation, the constraints include an external force constraint, a pressure actuation, and a length constraint. As discussed in \Cref{subsec:EquationsOfStaticEquilibrium}, constraints are gathered in $\boldsymbol{H}^T$ and $ \boldsymbol{\lambda}$ in \Cref{eq:equilibrium} ~\cite{coevoet_software_2017}. To find these two terms, we need to define the simulation boundary conditions.

\begin{figure}[ht]
    \centering
    \fontsize{7}{9}\selectfont
    \def\svgwidth{0.65\linewidth}
    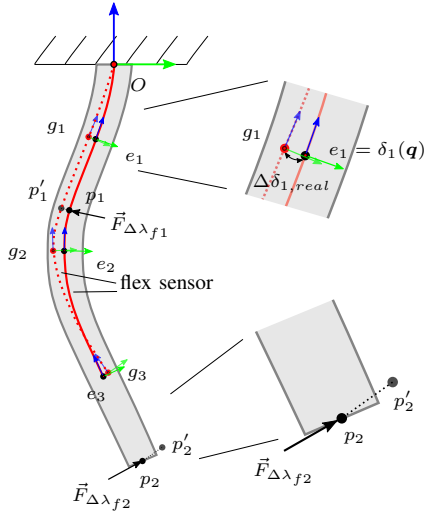

    \caption{\textbf{The constraints applied to the FEM model of the flexible strip.} External forces on the real device deform the flex sensor ({\color{red}\textbf{red}}), which is captured as a change in shape $\Delta \bm{\delta}_{real} = (g_{1}-e_{1}, \dots,  g_{n}-e_{n})^T$. SOFA finds the forces $\vec{F}_{\Delta\lambda fi}$ applied at points $p_{i}$ that minimize the distance of the effector points to the goal points. The forces bring about a change of shape of the whole device, including the points $p_{i}$ that move to their new positions $p'_{i}$. Note that the forces do not necessarily need to be applied on the section that is covered by the flex sensor, as illustrated here.}
    \label{fig:effector}
\end{figure}

The flexible strip is the simpler case, only requiring external force constraints. The external force constraint directly specifies the constraints in the task space. They are defined by a force magnitude $\boldsymbol{\lambda}$ and a direction at a certain point $p_{i}$ on the mesh encoded by $H_i$. 
The soft finger is additionally actuated by the pressurization of its internal chamber. Thus, the deformation of the soft finger is affected by the deformation of the chamber. The Lagrange multiplier $ \boldsymbol{\lambda}$ represents the pressure on the surface of the chamber, while $\boldsymbol{H}$ is related to the normal direction of the vertices of the mesh defining the surface of the chamber. Another major component affecting the dynamics is the combination of the metallic strip and the multi-tapped flex sensor embedded in the finger. 
Since the metallic strip and the flex sensor are fixed at both ends, they constrain the finger’s elongation on the front side, the one opposite to the chamber. We model this constraint using a length constraint with a sequence of fixed-length segments that follow along the length of the metallic strip. The Lagrange multiplier $\boldsymbol{\lambda}$, in this case, represents the force necessary to keep the length of each of the segments in the sequence constant. $\boldsymbol{H}$ is built using the constrained direction as if the sequence represented an inextensible cable~\cite{coevoet_optimization-based_2017}. 
\Cref{fig:effector} illustrates the external force constraints. Their relationship to solving an inverse problem for a given sensor configuration is illustrated in \Cref{fig:effector} and explained in the next subsection.

\subsection{Finding Constraining Forces Through Optimization}
\label{subsec:ForceOptimization}

We use inverse problem solving~\cite{coevoet_software_2017} to optimize for the generalized actuation forces $\bm{\lambda}$ that will minimize the differences between the simulated shape vectors and the sensed shape vectors. This means we search for actuation efforts that will best explain the observed deformation. We consider two types of forces: unknown external disturbances from pushing and known actuation by pressurized air in the chamber of the soft finger. In SOFA, the optimization is solved for each time step. Thus, in practice, the problem is solved by finding $\Delta\bm{\lambda}$ that corresponds to the difference in actuation forces that will minimize the error with regard to the change in shape for that time step.

$\delta_{e}:=\delta_{e}(\bm{q})$ is a function that maps the positions $\bm{q}$ of the FEM nodes to a pose of interest to the user, which is called a \emph{pose effector}. The pose in this case, could be cartesian coordinates only (3D) or include orientation as well (6D).
A displacement $\Delta\delta_{e,i}$ of an effector $i$ is composed of a positional and angular displacement (represented as angular velocity): $\Delta\delta_{e,i} = (\Delta x, \Delta y, \Delta z, \omega_{x}, \omega_{y}, \omega_{z})^T$. 
To find the relation between the change in applied forces and the change in shape in the simulation $\Delta \bm{\delta}_{e,sim}$ of a collection of position effectors, we can write:

\begin{equation}
\label{eq:DeltaR}
\Delta \bm{\delta}_{e,sim} = \bm{H}_{e} \bm{K}^{-1} \bm{H}_{f}^{T} \Delta \bm{\lambda} =  \bm{W}_{ef}\Delta \bm{\lambda}
\end{equation}
\label{eq:delta_p}
where $\bm{W}_{ef}:=\bm{H}_{e} \bm{K}^{-1} \bm{H}_{f}^{T} $ is the matrix that directly maps changes in actuation force to changes in the effector displacement (position and angular). This matrix can be obtained based on the constraints defined in \Cref{subsec:Constraints}.

Then the optimization problem can be formulated as follows:
\begin{equation}
\label{eq:Optimization}
\Delta\bm{\lambda}^{*} = \argmin_{\Delta \bm{\lambda}}{\lVert \bm{W}_{ef} \Delta \bm{\lambda} - \Delta \bm{\delta}_{real} \rVert ^2} 
\end{equation}
where the $\Delta \bm{\delta}_{real}=\bm{g}-\bm{e}=(g_{1}-e_{1}, \dots,  g_{n}-e_{n})^T=\Delta\bm{S}_{nom}$ are the real changes in the shape (see also \Cref{subsec:MappingRtoShape}). 
Furthermore, actuation efforts can also be taken into account during optimization by further constraints, such as
\begin{equation}
\label{eq:EqualityConstraints}
\Delta \bm{ \lambda}^{p} = \Delta \bm{P}_{input}
\end{equation}
\begin{equation}
\label{eq:InequalityConstraints}
\Delta \lambda_{i} \in [\Delta \lambda_{i, min}, \Delta \lambda_{i, max}] \;\forall i
\end{equation}

The equality constraints model that some of the efforts $\bm{\Delta\lambda}$ are known actuation of the pressurized chamber, as it is the case for the finger. The quadratic program (QP) is solved using a  Linear Complementarity Constraint Solver~\cite{coevoet_software_2017} and we obtain a value for all $\Delta\bm{\lambda}$ and, in particular, for the external forces $\Delta\bm{\lambda}^{f}$ due to the pushing.
\begin{equation}
\label{eq:ActuationVariables}
\Delta \bm{\lambda}=(\Delta \bm{\lambda}^{p}, \Delta \bm{\lambda}^{f})^T
\end{equation}
Therefore, the requirements (2) and (3) established in \Cref{sec:Intro} are met with a model capable of dealing with the solid mechanics, the sensing, and the actuation of the devices.

\section{Experimental Setup and Results}
\label{sec:Experiments}


\subsection{Data Collection}
\subsubsection{Flexible strip}
\label{subsubsec:FlexibleStrip}
We first tested our proposed method using the flexible TPU strip described in \Cref{sec:devices}.
We attach a series of 11 markers along the narrow side of the flexible strip for ground truth measurement. The flexible strip contains the sensor with an active length of \qty{80}{\milli\meter}. Note that the flexible strip is deliberately made to be longer (\qty{120}{\milli\meter}) than the sensor's active length, such that the multi-tapped sensor can only measure the shape of the strip corresponding to the first 8 markers (out of 11 total). The shape of the rest of the flexible strip will be inferred from the shape of the directly observable section of the strip using the FEM model without needing to retrain the neural network with this data.
The flexible strip is then attached rigidly at the base and deformed arbitrarily by hand to collect data. The training, validation, and test sets are recorded separately, where we deform the strip and log the resistance change and shape of the strip. For this device, we are interested in evaluating the accuracy of the shape reconstruction with the help of the solid mechanics model in SOFA.

\subsubsection{Soft finger}
Now we test our method on a pneumatically actuated soft finger. The active length of the finger is \qty{85}{\milli\meter}. 
Similar to the flexible strip, a series of markers are attached to the side of the finger for ground truth measurements.
The finger is actuated with different pressures ranging from \qtyrange{0}{1.8}{\kilo\pascal}, and an external force is applied on the finger by hand to collect training data. Note that in this case, the forces applied are not measured, and only the shape of the finger and corresponding resistance values are recorded.

The validation and test sets are recorded separately, where we use a robotic arm\footnote{Flexiv Rizon 4s} to apply an external force at a known location on the finger. The robotic arm has a 6-axis force-torque sensor to measure the force applied at the end-effector. 
The applied force is perpendicular to the contact surface, limiting the force to a single axis.
Therefore, the measured force is uniaxial along the normal axis of the contact surface.

We also use the validation set to identify the parameters for FEM (see \Cref{subsec:ParameterIdentificacion}). 
Actuation pressure for validation set and test set are \qty{1.5}{\kilo\pascal} and \qty{1.2}{\kilo\pascal}, respectively.
The external forces have range between \qtyrange{0}{5}{\newton} and are applied at different locations along the finger.
For each dataset, the data collection starts when the soft finger is fully pressurized, the robot arm approaches the soft finger and touches the finger's inner surface.
Then, the robot arm gradually increases and decreases the force magnitude while applying the force perpendicular to the contact surface.
After the contact force ramps down back to 0, the robot arm moves away from the finger and data collection stops.
We repeat this process with different pressures and different force magnitudes to construct the dataset.
For this device, we are interested in evaluating the accuracy of not only shape reconstruction but also the magnitude of the external force estimation.

\subsection{Resistances to Shape Regression}
The mapping from resistances to shape is modeled as described in \Cref{sec:FlexSensor}. We will now focus on specific details on the neural network training. Firstly, the data we collected was filtered and downsampled to 30 Hz to synchronize with the vision measurements.
In this dataset, unactuated neutral states dominate the data collection process, and deformed states are under-represented.
The training data is, therefore, resampled to improve the skewness of the dataset.
Nominal shape vectors are then computed from the measured marker positions using the FEM model. We used orientation change as the shape vector for the flexible strip and position changes for the soft finger. This difference comes from the different constraint settings in the two cases. The soft finger is constrained to have a constant length to ensure it bends under pressure actuation, while the flexible strip is not. Therefore, flexible strips are more sensitive to position errors. To avoid numerical instability, we used orientation change as the shape vector. 
Then we train the sensor model from this data.

For shape estimation of the sensor embedded in the flexible strip, we used a convolutional neural network with the input being a matrix containing the resistance vectors $\bm{R}_t$ stacked as follows: $[\bm{R}_{t-4}, \bm{R}_{t-3}, \bm{R}_{t-2}, \bm{R}_{t-1}, \bm{R}_{t}]^{\intercal}$, where $t$ refers to the timestep. 
The model consists of a convolutional layer of $32$ $3 \times 3$ filters with 0.5 dropout, flattened and fed to an output layer that predicts the relative position of each shape vector with respect to the previous one. 
The data collection procedure was run at \qty{30}{\hertz} for \qty{10}{\min}
to explore the state space and train this neural network.
However, in the case of the soft finger, the motion is more limited, and the state space that is being explored is limited as well. Therefore we chose a less expressive network with two fully connected layers of 32 and 16 neurons with 0.2 dropout. Input to the model is a vector of resistance measurements made at the current timestep. 
The data collection procedure was run at \qty{30}{\hertz} for \qty{15}{\min}
to explore the state space and train this neural network.

\subsection{Parametric Identification for the FEM model}
\label{subsec:ParameterIdentificacion}
For the flexible strip, the FEM model predicts the shape evolution beyond what the flex sensor can estimate. Since it is a purely geometrical problem, no material parameter identification was performed.  We chose the model parameters based on knowledge of material properties such that the deformation is feasible and the system converges. For the soft finger, the FEM model serves as a dynamic model for force estimation, and the parameters need to be calibrated experimentally, including the Poisson ratio and Young's Modulus.
In both cases, mesh density was chosen based on trial and error to match reality while maintaining a reasonable framerate for the simulation.
The material is modeled as quasi-incompressible, and the Poisson ratio is set to 0.45~\cite{zheng2019controllability}.
An approximate value of Young's Modulus is obtained by minimizing the position error with different actuation inputs ranging from \qtyrange{0}{2}{\kilo\pascal} with no external disturbances, which is determined to be \qty{1.37}{\mega\pascal}. 
Additionally, for force estimation, an empirical scaling factor of 1.5 is introduced to scale Young's Modulus and actuation pressure simultaneously to account for errors in the modeling process. This scaling factor is obtained by minimizing errors in force estimation in the validation set.

\begin{figure}[t]
    \centering
    \fontsize{7}{9}\selectfont
    \def\svgwidth{0.95\linewidth}
    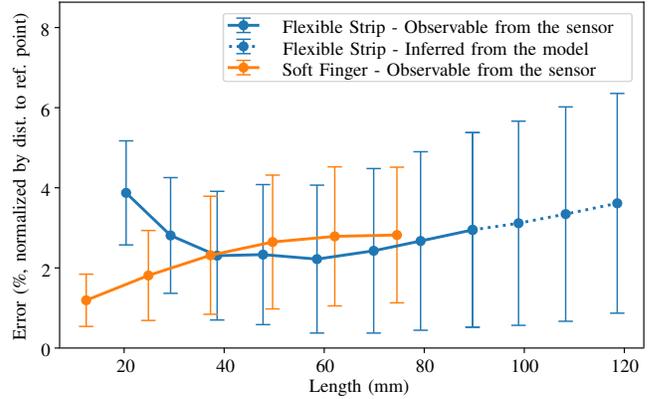
    \caption{\textbf{Reconstruction error for the flexible strip and soft finger.} On the x-axis is the distance along the length of the body from the fixed base.
    On the y-axis is the position error w.r.t vision-based ground truth measurement normalized by its distance to a fixed reference point at the base of the object.
    }
    \label{fig:Reconstruction_Error}
\end{figure}

\subsection{Experimental Results}
\label{subsec:Results}
\subsubsection{Shape Estimation of Flexible Strip}

As described in \Cref{subsec:ForceOptimization}, external forces are estimated through optimization to explain the observed deformation.
For the shape estimation of the flexible strip under deformation, a series of six external forces are arbitrarily applied at uniformly spaced locations along the strip.
The results of the shape estimation of the flexible strip are summarized in \Cref{fig:Reconstruction_Error} as reconstruction error statistics.
As mentioned previously in \Cref{subsubsec:FlexibleStrip}, only the first 8 points are covered by the sensor, and the rest are calculated by the FEM model based on the mechanics of the device.
The error is normalized by its distance to a fixed reference point at the base.
The error increases over the length of the device because the error accumulates over each marker. 
At the last point covered by the sensor, the average error is 2.95\% with respect to its distance to the reference point. 
At the end of the flexible strip, where the sensor is not present, the average error is 3.61\%  with respect to its distance to the reference point. 
The ability to make an estimation at the location, where the sensor is not present, demonstrates one of the advantages of the FEM model in comparison to a data-based model for extrapolating a system (requirement (2)).

\subsubsection{Shape and Force Estimation with the Soft Finger}

\begin{figure}[t]
    \centering
    \fontsize{7}{9}\selectfont
    \def\svgwidth{0.95\linewidth}
    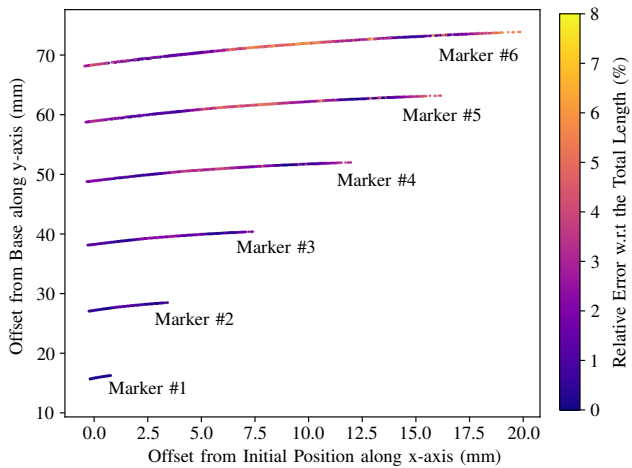
    \caption{\textbf{Error distributions in the estimation workspace of the soft finger.} The scattered points are the marker locations and the color scale represents the error magnitude. Markers are labeled according to the distance to the reference point}
    \label{fig:Error_Distribution_in_Workspace}
\end{figure}

 \Cref{fig:Reconstruction_Error} summarizes the shape reconstruction error for the soft finger in the test set. 
Similar to the flexible strip, the error increases over the length of the finger. 
At the furthest point from the base, the average error is 2.82\% with respect to its distance from the reference point.
Error distributions at different locations are illustrated in \Cref{fig:Error_Distribution_in_Workspace}. 

\begin{figure}[ht]
    \centering
    \fontsize{7}{9}\selectfont
    \def\svgwidth{0.95\linewidth}
    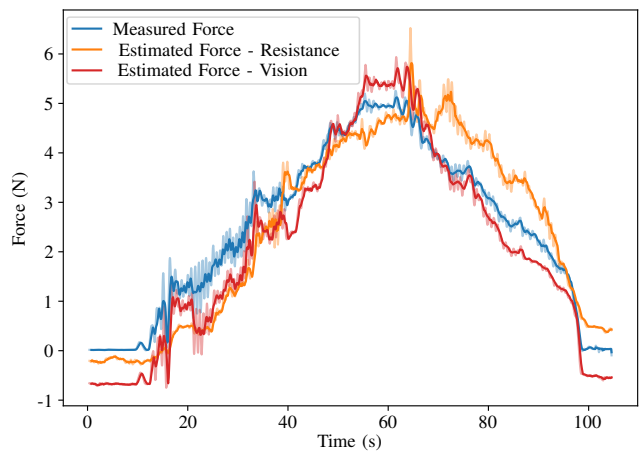
    \caption{\textbf{Measured and estimated forces for the soft pressurized finger.} The measured forces are obtained using a force-torque sensor behind a pointed tip, see \Cref{fig:demo_sys} (c).  The estimated forces use a FEM model and either a shape estimation from a multi-tap flex sensor or a visual shape measurement.
    }
    \label{fig:Estimated_Force_and_Measured_Force}
\end{figure}

The reconstructed shape of the finger is then used to estimate the external force. Results from a force estimation experiment are shown in \Cref{fig:Estimated_Force_and_Measured_Force}. 
We used shape measurement from both camera and resistive sensors to investigate how much error is introduced from the sensor mapping.
The estimated force using the camera follows the measured force better without the error of the additional mapping. 
The average relative errors are 10.18 \% and 11.82 \% with respect to the force range for vision and resistive estimations, respectively. 

\subsubsection{Discussion of the Experimental Results}
\label{subsec:Discussion}
Our position estimation reaches around 3\%  error with respect to the total length within the sensor coverage, which is a substantial improvement over the results reported in~\cite{thuruthel_soft_2019} with a flex sensor.
The errors are increasing as the measurement point moves away from the reference point. 
The errors are not consistent during the experiments, as shown in Fig. \ref{fig:Error_Distribution_in_Workspace}. 
The error reaches its maximum in the middle and furthest locations, which we believe is caused by the nonlinearity that our model fails to capture.
Apart from the nonlinearity, the dynamic motion of our experiment will also contribute to the model's error, which can explain the error in the middle.
Furthermore, the performance of our sensors is comparable to the cPDMS sensors Thuruthel et al. proposed without the need to fabricate a customized integrated sensor for individual robots.
Additionally, we were able to achieve a much larger range and a slightly better relative error for the magnitude of the force estimation.
The range of the force being estimated in our model-based force estimation is \qty{5}{\newton} with approximately 11\% relative error, and the range reported using a data-driven approach is \qty{0.5}{\newton} with 15.3 \% relative error~\cite{thuruthel_soft_2019}. 
In our experiments, the estimations are higher than the measurements after reaching the peak. We believe such behavior is related to hysteresis in the resistive sensor that is not adequately modeled by the neural network. 
The advantage of the model-based approach is the generalization ability to different object shapes and locations of action without training a new model. 
An advantage of the data-driven approach is the speed of inference. 
For example, the model in~\cite{thuruthel_soft_2019} was tested in real-time at around 10 Hz.
Our model for flexible strips was run in real-time at 30 Hz, but the structural complexity of the soft finger requires a finer mesh for accurate simulation, and therefore, the force estimation currenly runs at around \qty{5}{\hertz}.

\section{Conclusions and Future Work}
\label{sec:Conclusions}
In this work, we identified and addressed three key requirements for enabling proprioceptive shape and force sensing in soft robotics. The first is a technology that allows high-resolution shape sensing for capturing local deformations. The second is the use of a model that estimates forces based on the solid mechanics of the soft robotic device. The third is the ability to integrate and account for known actuation inputs in the model during force estimation. Based on these requirements, we proposed a multi-tapped resistive and modeling-based approach to expand the sensing capabilities of a widely available off-the-shelf resistive flex sensor.
We employed a neural network to interpret the nonlinear, coupled behavior of the multi-tap sensor readings. The resulting sensor signals informed an FEM-based model of the soft finger that also includes fluidic actuation. 
Besides the shape, we also estimated an external force acting on the body. The range of the external force was tested for \qtyrange{0}{5}{\newton} at arbitrary known locations. Our results show the potential of a multi-tapped resistive flex sensor when used with FEM-based modeling in a physical soft robotic gripper for shape reconstruction and force estimation. We believe this approach will bring us closer to proprioceptively grasping with soft grippers in unstructured, human-centric environments.

We propose for the future to further explore the potential of resistive proprioception. First, the sensor itself can be customized and fabricated with a built-in multi-tap interface, which allows for widespread adoption.
Second, we would like to optimize our FEM model for force estimation to achieve real-time performance, for instance, using a data-driven approach on the FEM model towards model order reduction~\cite{goury_fast_2018}.
Third, we would like to implement a numerical model of the multi-tapped sensor in SOFA so that we can directly map the measurement into the simulation and solve for the optimal solution.
Lastly, we would like to investigate the possibility of contact detection and thereby force localization through shape estimation.

\section*{Acknowledgements}
This work has been funded by Credit Suisse, Festo SE \& Co. KG, and Agencia Nacional de Investigación y Desarrollo (ANID) in Chile with the ``Fondecyt the Iniciación'' grant number 11230505. Barnabas Gavin Cangan was partially funded through a Swiss Government Excellence Fellowship by the Swiss Federal Commission for Scholarships.

\bibliographystyle{./IEEEtranBST/IEEEtran}
\bibliography{./IEEEtranBST/IEEEabrv,./references}

\end{document}